\documentclass[10pt,conference,compsocconf]{IEEEtran}
\ifCLASSOPTIONcompsoc
  \usepackage[nocompress]{cite}
\else
  \usepackage{cite}
\fi
\ifCLASSINFOpdf
  \usepackage[pdftex]{graphicx}
\else
  \usepackage[dvips]{graphicx}
\fi
\usepackage{amssymb}
\usepackage{amsmath}

\usepackage[linesnumbered,vlined,ruled]{algorithm2e}

\begin{document}
%
\title{Optimizing the Trade-off between\\
 Single-Stage and Two-Stage Deep Object Detectors\\ using Image Difficulty Prediction}

\author{\IEEEauthorblockN{Petru Soviany, Radu Tudor Ionescu}
\IEEEauthorblockA{Department of Computer Science\\
University of Bucharest, Romania\\
E-mails: petru.soviany@yahoo.com, raducu.ionescu@gmail.com}}

\maketitle

\begin{abstract}
There are mainly two types of state-of-the-art object detectors. On one hand, we have two-stage detectors, such as Faster R-CNN (Region-based Convolutional Neural Networks) or Mask R-CNN, that $(i)$ use a Region Proposal Network to generate regions of interests in the first stage and $(ii)$ send the region proposals down the pipeline for object classification and bounding-box regression. Such models reach the highest accuracy rates, but are typically slower. On the other hand, we have single-stage detectors, such as YOLO (You Only Look Once) and SSD (Singe Shot MultiBox Detector), that treat object detection as a simple regression problem by taking an input image and learning the class probabilities and bounding box coordinates. Such models reach lower accuracy rates, but are much faster than two-stage object detectors. In this paper, we propose to use an image difficulty predictor to achieve an optimal trade-off between accuracy and speed in object detection. The image difficulty predictor is applied on the test images to split them into easy versus hard images. Once separated, the easy images are sent to the faster single-stage detector, while the hard images are sent to the more accurate two-stage detector. Our experiments on PASCAL VOC 2007 show that using image difficulty compares favorably to a random split of the images. Our method is flexible, in that it allows to choose a desired threshold for splitting the images into easy versus hard.
\end{abstract}


%
\IEEEpeerreviewmaketitle

\section{Introduction}

Object detection, the task of predicting the location of an object along with its class in an image, is perhaps one of the most important problems in computer vision. Nowadays, there are mainly two types of state-of-the-art object detectors, as briefly discussed next. On one hand, we have two-stage detectors, such as Faster R-CNN (Region-based Convolutional Neural Networks)~\cite{Ren-NIPS-2015} or Mask R-CNN~\cite{He-ICCV-2017}, that $(i)$ use a Region Proposal Network (RPN) to generate regions of interests in the first stage and $(ii)$ send the region proposals down the pipeline for object classification and bounding-box regression. Such models reach the highest accuracy rates, but are typically slower. On the other hand, we have single-stage detectors, such as YOLO (You Only Look Once)~\cite{Redmon-CVPR-2016} and SSD (Singe Shot MultiBox Detector)~\cite{Liu-ECCV-2016}, that treat object detection as a simple regression problem by taking an input image and learning the class probabilities and bounding box coordinates. Such models reach lower accuracy rates, but are much faster than two-stage object detectors. In this context, finding a model that provides the optimal trade-off between accuracy and speed is not an easy task. Based on the principles of curriculum learning~\cite{Bengio-ICML-2009}, we hypothesize that using more complex (two-stage) object detectors for difficult images and less complex (single-stage) detectors for easy images will provide an optimal trade-off between accuracy and speed, without ever having to change anything about the object detectors. To test our hypothesis in practice, we employ a recent approach for image difficulty estimation introduced by Ionescu et al.~\cite{Ionescu-CVPR-2016}. The approach is based on training a deep neural network to regress on the difficulty scores produced by human annotators. In order to achieve a trade-off between accuracy and speed in object detection, we apply the image difficulty predictor on the test images to split them into easy versus hard (difficult) images. Once separated, the easy images are sent to the faster single-stage detector, while the hard images are sent to the more accurate two-stage detector. Our experiments on PASCAL VOC 2007~\cite{Pascal-VOC-2007} show that using image difficulty as a primary cue for splitting the test images compares favorably to a random split of the images. Moreover, our method is simple and has the advantage that allows to choose the desired trade-off on a continuous scale.

The paper is organized as follows. Recent related work on object detection is presented in Section~\ref{sec_RelatedWork}. Our methodology is described in Section~\ref{sec_Method}. The object detection experiments are presented in Section~\ref{sec_Experiments}. Finally, we draw our conclusions in Section~\ref{sec_Conclusion}.

\section{Related Work}
\label{sec_RelatedWork}

Although there are quite a few models for the object detection task available in the recent literature~\cite{Ren-NIPS-2015,He-ICCV-2017,Redmon-CVPR-2016,Liu-ECCV-2016,Howard-arXiv-2017}, it is difficult to pick one as the best model in terms of both accuracy and speed. Some~\cite{Ren-NIPS-2015,He-ICCV-2017} are more accurate and require a higher computational time, while others~\cite{Redmon-CVPR-2016,Liu-ECCV-2016,Howard-arXiv-2017} are much faster, but provide less accurate results. Hence, finding the optimal trade-off between accuracy and speed is not a trivial task. To our knowledge, the only work that studied the trade-off between accuracy and speed for deep object detection models is~\cite{Huang-CVPR-2017}. Huang et al.~\cite{Huang-CVPR-2017} have tested different configurations of deep object detection frameworks by changing various components and parameters in order to find optimal configurations for specific scenarios, e.g. deployment on mobile devices. Different from their approach, we treat the various object detection frameworks as black boxes. Instead of looking for certain configurations, we propose a framework that allows to set the trade-off between accuracy and speed on a continuous scale, by specifying the point of splitting the test images into easy versus hard, as desired. 

In the rest of this section, we provide a brief description of the most recent object detectors, in chronological order. Faster R-CNN~\cite{Ren-NIPS-2015} is a very accurate region-based deep detection model which improves Fast R-CNN~\cite{Girshick-ICCV-2015} by introducing the Region Proposal Networks. It uses a fully convolutional network that can predict object bounds at every location in order to solve the challenge of selecting the right regions. In the second stage, the regions proposed by the RPN are used as an input for the Fast R-CNN model, which will provide the final object detection results. 
On the other hand, SSD~\cite{Liu-ECCV-2016} is a single-shot detection method which uses a set of predefined boxes of different aspect ratios and scales in order to predict the presence of an object in a certain image. SSD does not include the traditional proposal generation and resampling stages, common for two-stage detectors such as Faster R-CNN, but it encapsulates all computations in a single network, thus being faster than the two-stage models. YOLO~\cite{Redmon-CVPR-2016} is another fast model, which treats the detection task as a regression problem. It uses a single neural network to predict the bounding boxes and the corresponding classes, taking the full image as an input. The fact that it does not use sliding window or region proposal techniques provides more contextual information about classes. YOLO works by dividing each image into a fixed grid, and for each grid location, it predicts a number of bounding boxes and a confidence for each bounding box. The confidence reflects the accuracy of the bounding box and whether the bounding box actually contains an object (regardless of class). YOLO also predicts the classification score for each box for every class in training. MobileNets~\cite{Howard-arXiv-2017} are a set of lightweight models that can be used for classification, detection and segmentation tasks. Although their accuracy is not as high as that of the state-of-the-art very deep models, they have the great advantage of being very fast and low on computational requirements, thus being suitable for mobile devices. MobileNets are built on depth-wise separable convolutions with a total of $28$ layers, and can be further parameterized in order to work even faster.
Mask R-CNN~\cite{He-ICCV-2017} is yet another model used in image detection and segmentation tasks, which extends the Faster R-CNN architecture. If Faster R-CNN has only two outputs, the bounding boxes and the corresponding classes, Mask R-CNN also provides, in parallel, the segmentation masks. An important missing piece of the Faster R-CNN model is a pixel alignment method. To address this problem, He et al.~\cite{He-ICCV-2017} propose a new layer (RoIAlign) that can correct the misalignments between the regions of interest and the extracted features.


\section{Methodology}
\label{sec_Method}

\begin{algorithm}[!th]
\caption{Easy-versus-Hard Object Detection\label{alg_easy_to_hard}}

\textbf{Input}: 

$I$ -- an input test image;

$D_{fast}$ -- a fast (single-stage) object detector;

$D_{slow}$ -- a slow (two-stage) object detector;

$P$ -- an image difficulty predictor;

$t$ -- a threshold for dividing images into easy or hard;

\BlankLine
\textbf{Computation}:

\If{$P(I) \leq t$}
{
	$B \leftarrow D_{fast}(I)$\;
}
\Else
{
	$B \leftarrow D_{slow}(I)$\;
}

\BlankLine
\textbf{Output}: 

$B$ -- the set of predicted bounding boxes.
\end{algorithm}

\begin{table*}[!th]
\caption{Mean Average Precision (mAP) and time comparison between MobileNet-SSD~\cite{Howard-arXiv-2017}, Faster R-CNN~\cite{Ren-NIPS-2015} and various combinations of the two object detectors on PASCAL VOC 2007. The test data is partitioned based on a random split (baseline) or an easy-versus-hard split given by the image difficulty predictor. For the random split, we report the average mAP over 5 runs to reduce bias. The times are measured on a computer with Intel Core i7 $2.5$ GHz CPU and $16$ GB of RAM.}
\begin{center}
\begin{tabular}{|l|c|c|c|c|c|}
\hline
                                        & \multicolumn{5}{|c|}{MobileNet-SSD (left) to Faster-RCNN (right)}\\
\cline{2-6}
                                        & $100\%-0\%$   & $75\%-25\%$   & $50\%-50\%$   & $25\%-75\%$   & $0\%-100\%$ \\
\hline
Random Split (mAP)	     	            & $0.6668$		& $0.6895$		& $0.7131$		& $0.7450$		& $0.7837$\\
Easy-versus-Hard Split (mAP)	        & $0.6668$		& $0.6981$		& $0.7431$		& $0.7640$		& $0.7837$\\
\hline
Image Difficulty Prediction Time (s)    & -             & $0.05$        & $0.05$        & $0.05$        & - \\
Object Detection Time (s)               & $0.07$		& $2.38$		& $4.08$		& $6.07$		& $7.74$\\
Total Time (s)                          & $0.07$		& $2.43$		& $4.13$		& $6.12$		& $7.74$\\
\hline
\end{tabular}
\end{center}
\label{Tab_FasterMobile_Results}
\end{table*}

Humans learn much better when the examples are not randomly presented, but organized in a meaningful order which illustrates gradually more complex concepts. This is essentially reflected in all the curricula taught in schooling systems around the world. Bengio et al.~\cite{Bengio-ICML-2009} have explored easy-to-hard strategies to train machine learning models, showing that machines can also benefit from learning by gradually adding more difficult examples. They introduced a general formulation of the easy-to-hard training strategies known as \emph{curriculum learning}. However, we can hypothesize that an \emph{easy-versus-hard} strategy can also be applied at test time in order to obtain an optimal trade-off between accuracy and processing speed. For example, if we have two types of machines (one that is simple and fast but less accurate, and one that is complex and slow but more accurate), we can devise a strategy in which the fast machine is fed with the easy test samples and the complex machine is fed with the difficult test samples. This kind of strategy will work as desired especially when the fast machine can reach an accuracy level that is close to the accuracy level of the complex machine for the easy test samples. Thus, the complex and slow machine will be used only when it really matters, i.e. when the examples are too difficult for the fast machine. The only question that remains is how to determine if an example is easy or hard in the first place. If we focus our interest on image data, the answer to this question is provided by the recent work of Ionescu et al.~\cite{Ionescu-CVPR-2016}, which shows that the difficulty level of an image (with respect to a visual search task) can be automatically predicted. With an image difficulty predictor at our disposal, we can test our hypothesis in the context of object detection from images. To obtain an optimal trade-off between accuracy and speed in object detection, we propose to employ a more complex (two-stage) object detector, e.g. Faster R-CNN~\cite{Ren-NIPS-2015}, for difficult test images and a less complex (single-stage) detector, e.g. SSD~\cite{Liu-ECCV-2016}, for easy test images. Our simple easy-versus-hard strategy is formally described in Algorithm~\ref{alg_easy_to_hard}. Since we apply this strategy at test time, the object detectors as well as the image difficulty predictor can be independently trained beforehand. This allows us to directly apply state-of-the-art pre-trained object detectors~\cite{Ren-NIPS-2015,Liu-ECCV-2016,Howard-arXiv-2017}, essentially as black boxes. On the other hand, we train our own image difficulty predictor as described below.

\noindent
{\bf Image difficulty predictor.}
We build our image difficulty prediction model based on CNN features and linear regression with $\nu$-Support Vector Regression ($\nu$-SVR)~\cite{Suykens-NPL-1999,taylor-Cristianini-cup-2004}. For a faster processing time, we consider a rather shallow pre-trained CNN architecture, namely VGG-f~\cite{Chatfield-BMVC-14}. The CNN model is trained on the ImageNet Large-Scale Visual Recognition Challenge (ILSVRC) benchmark~\cite{Russakovsky2015}.
We remove the last layer of the CNN model and use it to extract deep features from the fully-connected layer known as \emph{fc7}. The $4096$ CNN features extracted from each image are normalized using the $L_2$-norm. The normalized feature vectors are then used to train a $\nu$-SVR model to regress to the ground-truth difficulty scores provided by Ionescu et al.~\cite{Ionescu-CVPR-2016} for the PASCAL VOC 2012 data set~\cite{Pascal-VOC-2012}. We use the learned model as a continuous measure to automatically predict image difficulty. Our predictor attains a Kendall's $\tau$ correlation coefficient~\cite{upton-dict-stat-2008} of $0.441$ on the test set of Ionescu et al.~\cite{Ionescu-CVPR-2016}. We note that Ionescu et al.~\cite{Ionescu-CVPR-2016} obtain a higher Kendall's $\tau$ score ($0.472$) using a deeper CNN architecture~\cite{Simonyan-ICLR-14} along with VGG-f. However, we are interested in using an image difficulty predictor that is faster than all object detectors, even faster than MobileNets~\cite{Howard-arXiv-2017}, so we stick with the shallower VGG-f architecture, which reduces the computational overhead at test time.

\begin{figure*}[!th]

\begin{center}
\includegraphics[width=0.74\linewidth]{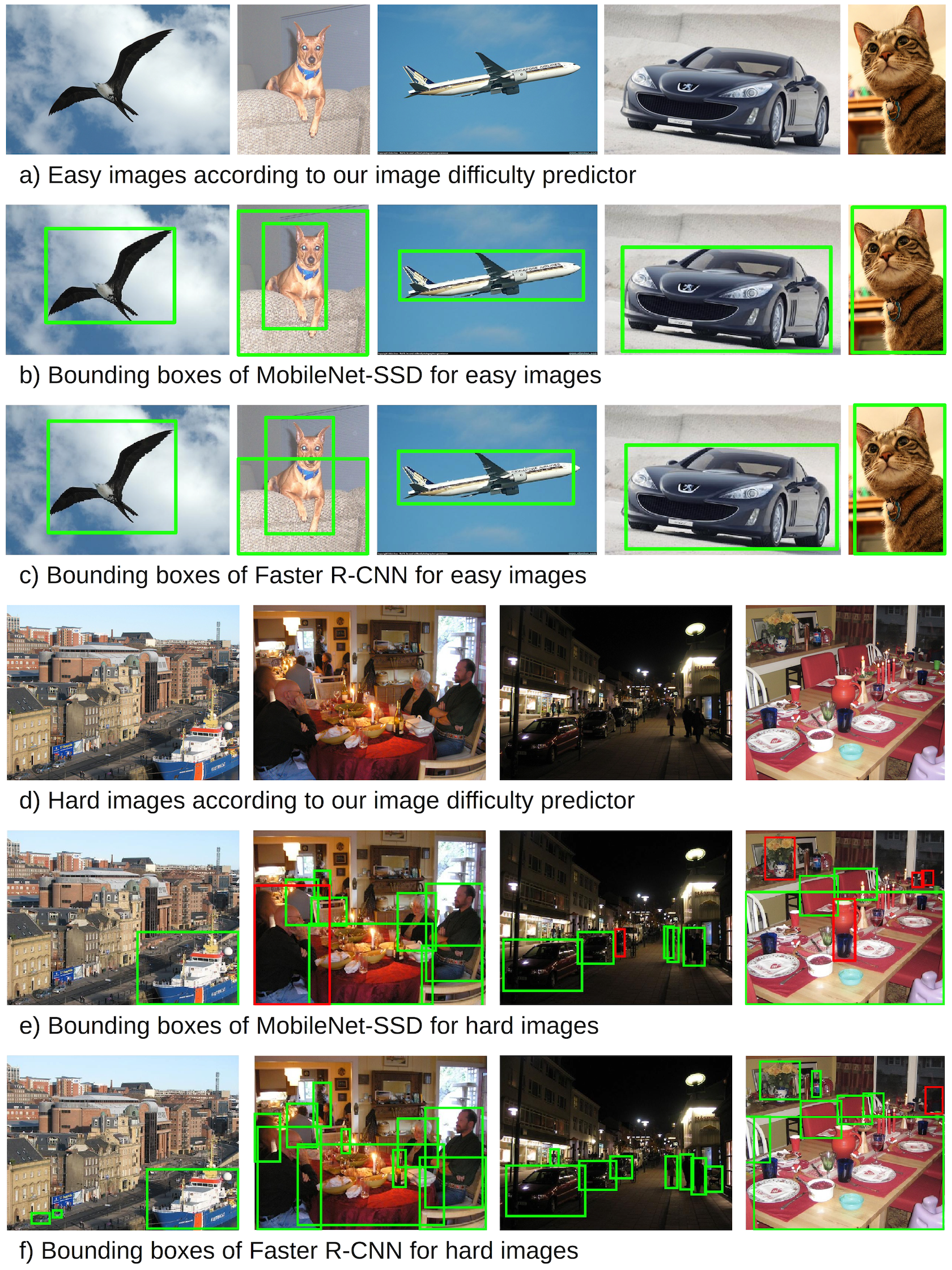}
\end{center}
\caption{Examples of easy (top three rows) and hard images (bottom three rows) from PASCAL VOC 2007 according to our image difficulty predictor. For each set of images, the bounding boxes predicted by the MobileNet-SSD~\cite{Howard-arXiv-2017} and the Faster R-CNN~\cite{Ren-NIPS-2015} detectors are also presented. The correctly predicted bounding boxes are shown in green, while the wrongly predicted bounding boxes are shown in red. Best viewed in color.}
\label{fig_easy_vs_hard}
\end{figure*}

\section{Experiments}
\label{sec_Experiments}

\subsection{Data Set}

We perform object detection experiments on the PASCAL VOC 2007 data set~\cite{Pascal-VOC-2007}, which consists of $9963$ images that contain $20$ object classes. The training and validation sets have roughly $2500$ images each, while the test set contains about $5000$ images.

\subsection{Evaluation Measure}

The performance of object detectors is typically evaluated using the mean Average Precision (mAP) over classes, which is based on the ranking of detection scores for each class~\cite{Pascal-VOC-2010}. For each object class, the Average Precision is given by the area under the precision-recall (PR) curve for the detected objects. The PR curve is constructed by first mapping each detected bounding box to the most-overlapping ground-truth bounding box, according to the Intersection over Union (IoU) measure, but only if the IoU is higher than $50\%$~\cite{Everingham-IJCV-2015}. Then, the detections are sorted in decreasing order of their scores. Precision and recall values are computed each time a new positive sample is recalled. The PR curve is given by plotting the precision and recall pairs as lower scored detections are progressively included.

\subsection{Models and Baselines}

We choose Faster R-CNN~\cite{Ren-NIPS-2015} based on the ResNet-101~\cite{He-CVPR-2016} architecture as our two-stage object detector that provides accurate bounding boxes. We set its confidence threshold to $0.6$. In the experiments, we use the pre-trained Faster R-CNN model available at {https://github.com/endernewton/tf-faster-rcnn}. 
We experiment with two single-shot detectors able to provide fast object detections, namely MobileNet-SSD~\cite{Howard-arXiv-2017} and SSD300~\cite{Liu-ECCV-2016}. We use the pre-trained MobileNet-SSD model available at {https://github.com/chuanqi305/MobileNet-SSD}. For SSD300, we use the model provided at {https://github.com/weiliu89/caffe/tree/ssd}, which is based on the VGG-16~\cite{Simonyan-ICLR-14} architecture. SSD300 takes input images of $300 \times 300$ pixels and performs the detection task in a single step. We also tried the SSD512 detector, but we did not find it interesting for our experiments, since its speed is a bit too high for a fast object detector ($1.57$ seconds per image). 

The main goal of the experiments is to compare two different strategies for splitting the images between the single-stage detector (MobileNet-SSD or SSD300) and the two-stage detector (Faster R-CNN). The first strategy is a baseline that splits the images randomly. To reduce the accuracy variation introduced by the random selection, we repeat the experiment for $5$ times and average the resulted mAP scores. We note that all standard deviations are lower than $0.5\%$. The second strategy is based on splitting the images into easy or hard, according to the difficulty scores assigned by our image difficulty predictor, as described in Section~\ref{sec_Method}.

\subsection{Results and Discussion}

Table~\ref{Tab_FasterMobile_Results} presents the mAP scores and the processing times of MobileNet-SSD~\cite{Howard-arXiv-2017}, Faster R-CNN~\cite{Ren-NIPS-2015} and several combinations of the two object detectors, on the PASCAL VOC 2007 data set. Different model combinations are obtained by varying the percentage of images processed by each detector. The table includes results starting with a $100\%-0\%$ split (equivalent with MobileNet-SSD~\cite{Howard-arXiv-2017} only), going through three intermediate splits ($75\%-25\%$, $50\%-50\%$, $25\%-75\%$) and ending with a $0\%-100\%$ split (equivalent with Faster R-CNN~\cite{Howard-arXiv-2017} only). In the same manner, Table~\ref{Tab_FasterSSD300_Results} shows the results for similar combinations of SSD300~\cite{Liu-ECCV-2016} and Faster R-CNN~\cite{Ren-NIPS-2015}. 

We first analyze the mAP scores and the processing time of the three individual object detectors, namely Faster R-CNN~\cite{Ren-NIPS-2015}, MobileNet-SSD~\cite{Howard-arXiv-2017} and SSD300~\cite{Liu-ECCV-2016}. Faster R-CNN reaches a mAP score of $0.7837$ in about $7.74$ seconds per image, while SSD300 reaches a mAP score of $0.69$ in $0.56$ seconds per image. MobileNet-SDD is even faster, attaining a mAP score of $0.6668$ in just $0.07$ seconds per image. We hereby note that we also considered the SSD512 object detector, but its results (a mAP score of $0.7046$ in $1.57$ seconds per image) did not convince us to include it the evaluation.

\begin{table*}[!th]
\caption{Mean Average Precision (mAP) and time comparison between SSD300~\cite{Liu-ECCV-2016}, Faster R-CNN~\cite{Ren-NIPS-2015} and various combinations of the two object detectors on PASCAL VOC 2007. The test data is partitioned based on a random split (baseline) or an easy-versus-hard split given by the image difficulty predictor. For the random split, we report the average mAP over 5 runs to reduce bias. The times are measured on a computer with Intel Core i7 $2.5$ GHz CPU and $16$ GB of RAM.}
\begin{center}
\begin{tabular}{|l|c|c|c|c|c|}
\hline
                                        & \multicolumn{5}{|c|}{SSD300 (left) to Faster-RCNN (right)}\\
\cline{2-6}
                                        & $100\%-0\%$   & $75\%-25\%$   & $50\%-50\%$   & $25\%-75\%$   & $0\%-100\%$ \\
\hline
Random Split (mAP)	     	            & $0.6900$		& $0.7003$		& $0.7178$		& $0.7561$		& $0.7837$\\
Easy-versus-Hard Split (mAP)	        & $0.6900$		& $0.7117$		& $0.7513$		& $0.7732$		& $0.7837$\\
\hline
Image Difficulty Prediction Time (s)    & -             & $0.05$        & $0.05$        & $0.05$        & - \\
Object Detection Time (s)               & $0.56$		& $2.46$		& $4.33$		& $6.12$		& $7.74$\\
Total Time (s)                          & $0.56$		& $2.49$		& $4.38$		& $6.17$		& $7.74$\\
\hline
\end{tabular}
\end{center}
\label{Tab_FasterSSD300_Results}
\end{table*}

We next analyze the average processing times per image of the various model combinations. As expected, the time improves by about $21\%$ when running MobileNet-SSD on $25\%$ of the test set and Faster R-CNN on the rest of $75\%$. On the $50\%-50\%$ split, the processing time is nearly $47\%$ shorter than processing the entire test set with Faster R-CNN only ($0\%-100\%$ split). On the $75\%-25\%$ split, the processing time further improves by $69\%$. As SSD300 is slower than MobileNet-SSD, the time improvements are close, but not as high. The improvements in terms of time are $20\%$ for the $25\%-75\%$ split, $44\%$ for the $50\%-50\%$ split, and $68\%$ for the $75\%-25\%$ split. We note that unlike the random splitting strategy, the easy-versus-hard splitting strategy requires additional processing time for computing the difficulty scores. The image difficulty predictor runs in about $0.05$ seconds per image. However, the extra time required by the difficulty predictor is almost insignificant with respect to total time required by the various combinations of object detectors. For instance, in the $50\%-50\%$ split with MobileNet-SSD and Faster R-CNN, the difficulty predictor accounts for roughly $1\%$ of the total processing time ($0.05$ out of $4.13$ seconds per image).

Regarding the two strategies for combining object detectors, the empirical results indicate that the easy-versus-hard splitting strategy gives better performance for all model combinations. The highest differences between the two strategies can be observed for the $50\%-50\%$ split. When using MobileNet-SSD for the easy images (Table~\ref{Tab_FasterMobile_Results}), our strategy gives a performance boost of $3\%$ (from $0.7131$ to $0.7431$) over the random splitting strategy. However, the mAP of the MobileNet-SSD and Faster R-CNN combination is $4.06\%$ under the mAP of the standalone Faster R-CNN. When using SSD300 for the easy images (Table~~\ref{Tab_FasterSSD300_Results}), our strategy gives a performance boost of $3.35\%$ (from $0.7178$ to $0.7513$) over the baseline strategy. This time, the mAP of the SSD300 and Faster R-CNN combination is $3.24\%$ under the mAP of the standalone Faster R-CNN, although the processing time is reduced by almost half.

To understand why our easy-versus-hard splitting strategy gives better results than the random splitting strategy, we randomly select a few easy examples and a few difficult examples from the PASCAL VOC 2007 data set, and we display them in Figure~\ref{fig_easy_vs_hard} along with the bounding boxes predicted by the MobileNet-SSD and the Faster R-CNN object detectors. On the easy images, the bounding boxes of the two detectors are almost identical. There is however an observable difference for the image that depicts a dog sitting on a sofa (second image from the left, on the second row in Figure~\ref{fig_easy_vs_hard}), as the bounding box provided by MobileNet-SSD for the sofa includes too much of the background. Nevertheless, we can perceive a lot more differences between MobileNet-SSD and Faster R-CNN on the hard images. In the left-most hard image, Faster R-CNN is able to detect two small cars, besides the large vessel. In the second and the third images, MobileNet-SSD misses some of the smaller objects and also provides wrong bounding boxes. In the right-most hard image, MobileNet-SSD misses some of the objects and provides a wrong label (TV/monitor) for the potted plant sitting behind the table. We thus conclude that the difference between MobileNet-SSD and Faster R-CNN is less noticeable on the easy images than on the hard images. This could explain why our easy-versus-hard splitting strategy is effective in choosing an optimal trade-off between accuracy and speed.

\section{Conclusion}
\label{sec_Conclusion}

In this paper, we have presented an easy-versus-hard strategy to obtain an optimal trade-off between accuracy and speed in object detection from images. Our strategy is based on dispatching the test images according to their difficulty (easy or hard) either to a fast and less accurate single-shot detector or to a slow and more accurate two-stage detector. We have conducted experiments using state-of-the-art object detectors such as SSD300~\cite{Liu-ECCV-2016} or Faster R-CNN~\cite{Ren-NIPS-2015} on the PASCAL VOC 2007~\cite{Pascal-VOC-2007} data set. The empirical results indicate that using image difficulty as a primary cue for splitting the test images compares favorably to a random split of the images. Furthermore, our approach is simple and easy to use by anyone in practice.

In future work, we aim to study and experiment with other strategies for dispatching the images to an appropriate object detector. We also aim to investigate whether training object detectors to specifically deal with easy or hard image samples can help to further improve our results. 

\ifCLASSOPTIONcompsoc
\section*{Acknowledgments}
The work of Petru Soviany was supported through project grant PN-III-P2-2.1-PED-2016-1842. The work of Radu Tudor Ionescu was supported through project grant PN-III-P1-1.1-PD-2016-0787.
\else
\section*{Acknowledgments}
The work of Petru Soviany was supported through project grant PN-III-P2-2.1-PED-2016-1842. The work of Radu Tudor Ionescu was supported through project grant PN-III-P1-1.1-PD-2016-0787.
\fi



\bibliographystyle{IEEEtran}
\bibliography{IEEEabrv,references}

\end{document}